\documentclass[letterpaper]{article} 
\usepackage[submission]{aaai24}  
\usepackage{times}  
\usepackage{helvet}  
\usepackage{courier}  
\usepackage[hyphens]{url}  
\usepackage{graphicx} 
\usepackage{natbib}  
\usepackage{caption} 
\usepackage[ruled, vlined]{algorithm2e}
\usepackage{pifont}
\usepackage{algorithmic}
\usepackage{makecell}
\usepackage{amsmath}
\usepackage{multirow}
\usepackage{multicol}
\usepackage{float}
\usepackage{amssymb}
\usepackage{newfloat}
\usepackage{listings}
\usepackage{subcaption}
\usepackage{dashrule}
\usepackage{booktabs}

\usepackage{newfloat}
\usepackage{listings}
\DeclareCaptionStyle{ruled}{labelfont=normalfont,labelsep=colon,strut=off} 
\floatstyle{ruled}
\newfloat{listing}{tb}{lst}{}
\floatname{listing}{Listing}
\title{FedDAT: An Approach for Foundation Model Finetuning in Multi-Modal Heterogeneous Federated Learning}
\author{
   Haokun Chen\textsuperscript{\rm 1,}\textsuperscript{\rm 2} \quad
   Yao Zhang\textsuperscript{\rm 1} \quad
   Denis Krompass\textsuperscript{\rm 2} \quad 
   Jindong Gu\textsuperscript{\rm 3} \quad
   Volker Tresp\textsuperscript{\rm 1,}\textsuperscript{\rm 4}
}
\affiliations{
\textsuperscript{\rm 1} Ludwig Maximilian University of Munich \quad
\textsuperscript{\rm 2} Siemens Technology  \\
\textsuperscript{\rm 3} University of Oxford \quad
\textsuperscript{\rm 4} Munich Center for Machine Learning \\
\tt{\small \{haokun.chen, denis.krompass\}@siemens.com, yzhang@dbs.ifi.lmu.de \\ jindong.gu@outlook.com, volker.tresp@lmu.de} 
}

\usepackage{bibentry}

\begin{document}

\maketitle

\begin{abstract}
Recently, foundation models have exhibited remarkable advancements in multi-modal learning. These models, equipped with millions (or billions) of parameters, typically require a substantial amount of data for finetuning. However, collecting and centralizing training data from diverse sectors becomes challenging due to distinct privacy regulations. Federated Learning (FL) emerges as a promising solution, enabling multiple clients to collaboratively train neural networks without centralizing their local data. To alleviate client computation burdens and communication overheads, previous works have adapted Parameter-efficient Finetuning (PEFT) methods for FL. Hereby, only a small fraction of the model parameters are optimized and communicated during federated communications. Nevertheless, most previous works have focused on a single modality and neglected one common phenomenon, i.e., the presence of data heterogeneity across the clients. Therefore, in this work, we propose a finetuning framework tailored to heterogeneous multi-modal FL, called Federated Dual-Aadapter Teacher (FedDAT). Specifically, our approach leverages a Dual-Adapter Teacher (DAT) to address data heterogeneity by regularizing the client local updates and applying Mutual Knowledge Distillation (MKD) for an efficient knowledge transfer. FedDAT is the first approach that enables an efficient distributed finetuning of foundation models for a variety of heterogeneous Vision-Language tasks. To demonstrate its effectiveness, we conduct extensive experiments on four multi-modality FL benchmarks with different types of data heterogeneity, where FedDAT substantially outperforms the existing centralized PEFT methods adapted for FL.
\end{abstract}


\section{Introduction}
Recent works have shown the power of foundation models with millions (billions) of parameters \cite{zhou2023comprehensive, du2022survey}. These models, represented by Transfomers \cite{vaswani2017attention}, achieve promising results when finetuned for real-world multi-modal tasks, including Visual Question Answering (VQA) \cite{VQAabstract}, Visual Commonsense Reasoning (VCR) \cite{zellers2019vcr}, etc. To improve the generalization ability of the foundation models, a substantial amount of data from diverse sectors and application scenarios is typically required for extensive finetuning. However, it becomes challenging to aggregate all training data and perform centralized model finetuning. For instance, collecting data from different clinical centers across multiple countries becomes infeasible due to distinct privacy regulations, such as GDPR in the EU and PDPA in Singapore.

To address this problem, Federated Learning (FL) emerges as a promising solution, which allows a shared model to be collaboratively optimized using decentralized data sources. In the classical FL approaches, e.g., FedAvg \cite{mcmahan2017communication}, the central server obtains the model by iteratively averaging the optimized model weights uploaded from the active clients. FL offers several advantages, including improved efficiency in client-server communication and enhanced data confidentiality, as it eliminates the need for direct access to the client's local dataset. FL provides promising solutions for various application areas, such as healthcare \cite{sheller2020federated} and industry \cite{liu2020deep}, where data privacy is crucial.

\begin{figure}[t]
\includegraphics[scale=0.26]{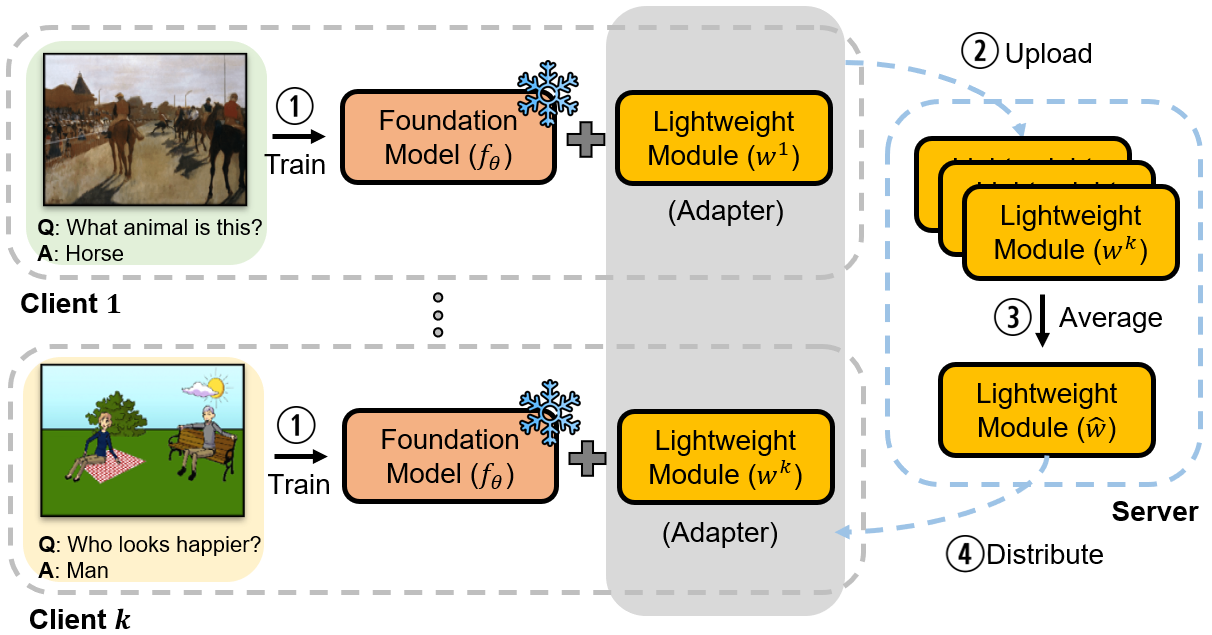} 
\vspace{-10pt}
\centering
\caption{Schematic illustration of the training procedure for Visual Question Answering (VQA) in Federated Learning.}
\vspace{-10pt}
\label{fig:fedvqa} 
\end{figure}

Despite its promising prospects, traditional FL is unsuitable for finetuning the entire foundation model. The optimization and transmission of billions of parameters would impose significant client computation burdens and substantial communication overheads. To overcome this challenge, parameter-efficient finetuning (PEFT) methods provide a possible solution, where only a small fraction of the model parameters is optimized and communicated during FL. 

Existing works have predominantly explored a basic combination of centralized PEFT algorithms and FedAvg. For instance, some approaches focus on training and communicating only the tiny adaptation modules (adapter) \cite{houlsby2019parameter, su2022cross} or a small amount of trainable input tokens \cite{guo2022promptfl, guo2023pfedprompt}. However, these investigations are limited to single modality scenarios, where only visual or textual tasks are considered. Most importantly, none of these works address the problem of data heterogeneity, in which the data of different clients are not independent and identically distributed (\emph{non-IID}). Data heterogeneity may lead to model drifts during the client local update, as well as an unstable and sub-optimal convergence of the aggregated server model \cite{li2020federatedchallenge, mendieta2022local}. Therefore, in this paper, we propose Federated Dual-Adapter Teacher (\texttt{FedDAT}), as the first framework to address this challenging yet practical problem, PEFT of foundation models for multi-modal (Vision-Language) heterogeneous FL.


\texttt{FedDAT} incorporates a global adapter in the foundation model, which is optimized and transmitted during federated communications. \texttt{FedDAT} utilizes a Dual-Adapter Teacher (\emph{DAT}) module, comprising two parallel adapters: one is a copy of the global adapter, kept frozen, while the other is locally optimized at each client. This configuration enables the local adapter to capture client-specific knowledge, which serves to regularize the global adapter and address data heterogeneity. Meanwhile, the frozen adapter preserves client-agnostic knowledge, thereby mitigating the catastrophic forgetting of the global adapter during knowledge transfer. To prevent overfitting of \emph{DAT} to the limited client local dataset, we implement Mutual Knowledge Distillation (\emph{MKD}) between \emph{DAT} and the global adapter. This mechanism ensures efficient knowledge transfer while maintaining the generalization ability of both modules. 


The proposed method \texttt{FedDAT} achieves state-of-the-art results on four multi-modality benchmarks that include a variety of Vision-Language (VL) tasks with data heterogeneity. Our contributions can be summarized as follows:

\begin{itemize}
\item We propose a novel method \texttt{FedDAT} for multi-modal heterogeneous FL, which is the first FL framework addressing distributed PEFT of foundation models for Vision-Language tasks. 

\item We conduct comprehensive experiments on four heterogeneous FL benchmarks with a variety of Vision-Language tasks. The results demonstrate that \texttt{FedDAT} achieves SOTA results, indicating better convergence rate and scalability compared to existing PEFT methods.

\end{itemize}

\section{Related Work}
\textbf{Parameter-Efficient Finetuning (PEFT) for Federated Learning:} 
PEFT has been well studied in centralized machine learning \cite{houlsby2019parameter, liu2022few, sung2022vl}, while its application on FL remains under-explored. Most of the prior work rudimentarily adapted PEFT for FL and focused on single-modal tasks: 

(1) Image classification: \cite{chen2022fedtune, sun2022exploring} evaluate the existing PEFT baselines combined with FL, while \cite{guo2022promptfl, guo2023pfedprompt, li2023visual, lu2023fedclip} finetune the CLIP model \cite{radford2021learning} via tuning and communicating only small amount of learnable (personalized) prompts. \cite{su2022cross} addresses the problem of heterogeneous client images by injecting lightweight adaptation modules (adapters) \cite{houlsby2019parameter}. \cite{yang2023exploring} explores the possibility of finetuning generative foundation models (diffusion models) \cite{dhariwal2021diffusion} via FL. 

(2) Language tasks: \cite{yu2023federated} requires public server dataset and optimize adapter for few-shot finetuning of BERT-like language models \cite{devlin2018bert}. \cite{zhang2023towards} builds a distributed instruction tuning \cite{wei2021finetuned} datasets and finetunes the language model via Low-Rank Adaptation (LoRA) \cite{hu2021lora}. \cite{zhuang2023foundation} systematically analyzes the challenges of finetuning large language models in FL. 

\cite{yu2023multimodal} is the first to analyze the situation of having multi-modal client datasets and conducts contrastive representation learning. However, the visual data and the language data are processed by separate networks, i.e., no Vision-Language Foundation Model is involved. In this work, we focus on the under-explored PEFT for large-scale vision-language models in FL and address the problem of client local datasets with heterogeneity in both vision and/or language modality.

\textbf{Vision-Language Foundation Model: } 
Vision-Language foundation models have significantly advanced the Vision-Language tasks \cite{VQAabstract, zellers2019vcr, DBLP:conf/acl/SuhrZZZBA19, DBLP:journals/corr/abs-1901-06706}. Based on the perspective of intra-modality data handling, there are two types of mainstream Vision-Language Foundation model structures: (1) Single-stream Vision-Language Foundation models \cite{DBLP:journals/corr/abs-1908-03557, DBLP:conf/eccv/ChenLYK0G0020, DBLP:conf/eccv/Li0LZHZWH0WCG20, DBLP:conf/iclr/SuZCLLWD20, DBLP:conf/icml/KimSK21, DBLP:conf/cvpr/SinghHGCGRK22}, which directly fuse the initial language/visual representation by using the joint cross-modal encoder at the initial state, and (2) Dual-stream Vision-Language foundation models \cite{DBLP:conf/nips/LuBPL19, DBLP:conf/emnlp/TanB19, DBLP:conf/nips/LiSGJXH21, huo2021wenlan}, which separately apply the intra-modality processing to two modalities along with a shared cross-modal encoder. To showcase the applicability of our proposed \texttt{FedDAT} to a wide range of Vision-Language foundation models, we carefully select ViLT \cite{DBLP:conf/icml/KimSK21} as a representative single-stream Vision-Language foundation model, and ALBEF \cite{DBLP:conf/nips/LiSGJXH21} as a representative dual-stream Vision-Language foundation model. By employing these diverse models, we effectively demonstrate the versatility and robustness of \texttt{FedDAT} in Vision-Language learning.

\section{Methodology}

\subsection{Problem Statement}

In this work, we address a heterogeneous FL problem setting with $K$ clients: Each client $k$ owns its private multi-modal dataset $D^k$, containing data from visual modality (images) and textual modality (texts). Specifically, we focus on the vision-language tasks and take Visual Question Answering (VQA) as an example. Hereby, the local dataset $D^k$ can be further decomposed into $N_k$ image-question-answer triplets $\{(v^k_i, q^k_i, a^k_i) | i \in \{1, ..., N_k\}\}$. We assume that the marginal distribution of $v^k_i$ and/or $q^k_i, a^k_i$ varies across the clients, i.e., there exists data heterogeneity in the visual space and/or in the textual space. We define the answer pool $A^k = \{a^k_1,...,a^k_{C^k}\}$ with $C^k$ ground-truth answers for client $k$ and define our task as a $C^k$-way classification problem following \cite{VQAabstract}. Note that the size of the answer pool could be different for different clients. The objective of FL is to collaboratively finetune one \emph{global} foundation model $f_{\theta}$ in a parameter-efficient manner (PEFT) within a pre-defined communication budget, which produces promising results on all client's local data.

\subsection{PEFT Method: Adapter}
\label{sec:baselines}
In this section, we introduce a traditional parameter-efficient finetuning (PEFT) method, i.e., Adapter \cite{houlsby2019parameter}, adjusted for FL applications. Here, we adopt the foundation models with common Transformer architecture \cite{vaswani2017attention} consisting of multiple repeated Transformer blocks. Specifically, each block contains a self-attention sub-layer, a fully connected feed-forward network (FFN), and residual connections around the sub-layers followed by layer normalization. 

Adapter is a bottleneck network consisting of a down-sample linear layer $W_{down} \in \mathbb{R}^{d \times r}$ and an up-sampling linear layer $W_{up} \in \mathbb{R}^{r \times d}$, where $r$ denotes the down-sampled dimension ($r<d$). A nonlinear activation function $\phi(\cdot)$, such as ReLU, is inserted in between. The adapter is injected after the FFN of each Transformer block and its computation can be formulated as

\vspace{-8pt}
\begin{equation}
h' = h + \phi(hW_{down})W_{up},    
\label{eq:adapter}
\end{equation}

where $h$ is the normalized output of FFN.

\begin{table}[b]
\vspace{-5pt}
\centering
\small
\setlength{\tabcolsep}{1.1mm}{
\begin{tabular}{c|cccccc|c}
\toprule
\multirow{2}{*}{Method} & \multicolumn{7}{c}{DomainNet}   \\
\cline{2-8}
~ & C & I & P & Q & R & S & avg \\
\hline
$clf$-L & 72.43 & 36.13 & \underline{86.35} & 55.70 & 74.07 & \underline{74.70} & 66.56\\
$Adapter$-L & 76.05 & 36.93 & \underline{88.03} & 72.40 & 66.53 & \underline{78.74} & 69.78\\
$clf$ & 80.80 & 44.61 & \underline{83.47} & 60.10 & 84.21 & \underline{71.69} & 70.81 \\
$Adapter$ & 88.59 & 50.95 & \underline{87.12} & 76.00 & 84.99 & \underline{74.08} & \textbf{76.96}\\
\bottomrule
\end{tabular}
\vspace{-0.3em}
\caption{Evaluation results of ViT optimized with different algorithms on federated image classification benchmark (DomainNet). "$L$" indicates independent client training, i.e., no federated communication involved.}
\label{tab:motiv}
}
\vspace{-0.6em}
\end{table}

\subsection{Recap: Federated Averaging}
In this section, we formally describe the combination of the conventional federated learning algorithm, FedAvg \cite{mcmahan2017communication}, and the centralized PEFT algorithm, i.e., Adapter. Before the client-server communication starts, we prepare the \emph{same} pre-trained foundation model $f_{\theta}$ for different clients. Afterwards, the server randomly initializes the parameter $w$ of the learnable lightweight module, which are the weight matrices of the linear layers $W_{down}$ and $W_{up}$ in the adapters. $w$ is then distributed to all clients for communication and local optimization. We illustrate the procedure of one communication round in the following. 

As shown in Figure \ref{fig:fedvqa}, each active client $k$ first execute local training to optimize the light-wegiht module $w^k$ combined with the \emph{frozen} foundation model $f_{\theta}$ (\ding{192}) in parallel, where the following loss $L_k$ is minimized:
\begin{equation}
    L_k(w^k) = \frac{1}{N_k} \sum_{i=1}^{N_k} \mathcal{L}(y_i, f_{\theta \cup w^k}(x_i)),
\end{equation}

where $y_i$ is the ground-truth label of input data $x_i$, and $\mathcal{L}$ is the loss function, e.g., Cross-Entropy for classification tasks. After the local updates, the central server aggregates $\{w^k|1 \leq k \leq K\}$, uploaded (\ding{193}) by all active clients, and executes a parameter aggregation (\ding{194}):
\begin{equation}
    \hat{w} \gets \frac{1}{\sum_{k=1}^K N_k} \sum_{k=1}^{K} N_k \cdot w^k.
    \label{eq:fedavg}
\end{equation}

Finally, the aggregated weight $\hat{w}$ will be distributed (\ding{195}) to the active clients for optimization in the next communication round. Note that after exhausting all communication budgets, the global model $f_{\theta \cup w}$ is deployed for the testing.

\subsection{Motivational Case Study}
\label{sec:motivstudy}
To motivate the architecture design of \texttt{FedDAT}, we present an empirical analysis to address the following research question: \emph{Which type of knowledge is more crucial for optimizing a promising ML model in heterogeneous FL, client-specific or client-agnostic?} Therefore, we follow the experiment design proposed in \cite{tan2022federated}. Specifically, we take the down-sampled version of DomainNet \cite{peng2019moment}, which is an image classification benchmark and contains data from 6 different styles: Clipart (C), Infograph (I), Painting (P), Quickdraw (Q), Real (R), and Sketch (S). By assigning data from one style to each client, we simulate data heterogeneity in the feature space across different clients. We finetune the foundation model, i.e., ViT \cite{dosovitskiy2020image}, with different PEFT methods via FL.

In Table \ref{tab:motiv}, we provide the results of finetuning the classification head ($clf$) and finetuning with $Adapter$. We also display the performance of client local finetuning ($L$), i.e., no federated communication involved. We conclude three observations from the results: (1) $Adapter$ is an effective PEFT method in both federated setting and independent finetuning setting compared with $clf$, providing an average performance increase of $3.22\%$ and $6.15\%$, respectively. (2) Collaborative training via FL, i.e., finetuning a \emph{client-agnostic} foundation model, generally outperforms local independent finetuning. This can be observed by comparing the average accuracy of models with and without "L". (3) \emph{Client-specific} classification head and adapters show benefits on certain clients (marked with underlines), i.e., clients with Painting (P) and Sketch (S) data and optimized independently. We assume this is due to the large distribution shift in the feature space across different clients' local data, given their different image appearances. This phenomenon answers the previous research question: Both \emph{client-specific} and \emph{client-agnostic} knowledge are crucial and should not be forgotten during federated communication. These observations motivate the proposed method and serve as evidence for its promising applicability and effectiveness.

\subsection{Proposed Method}

In this section, we introduce the proposed method Federated Dual-Adapter Teacher (\texttt{FedDAT}). As shown in Algorithm \ref{algo:FedDAT}, the training process of \texttt{FedDAT} can be divided into two functions, which will be introduced in the following:

\subsubsection{Server Update:} At the beginning of the training, the server initializes a shared adapter $A_s$. In each communication round, all active clients receive $A_s$ and conduct $Client \ Update$ in parallel. Subsequently, the server aggregates and averages the optimized parameters $\{A_s^k | 1 \leq k \leq K\}$ uploaded from all clients, which will be used as the initialization of $A_s$ for the next communication round.


\subsubsection{Client Update:} The client local update comprises 2 main components, which will be introduced in the following: 

\begin{figure}[t]
\begin{subfigure}[b]{0.25\textwidth}
    \includegraphics[scale=0.225]{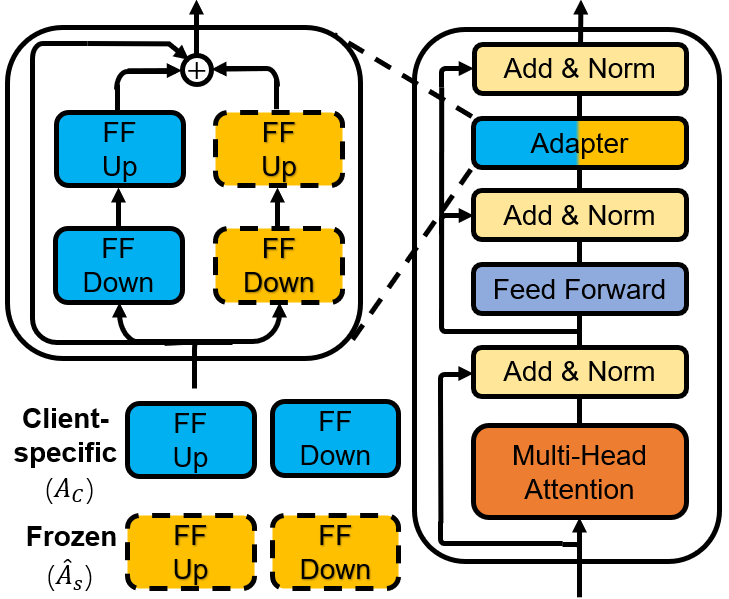} 
    \flushleft 
    \caption{\emph{DAT} Architecture Design}
    \label{fig:FedDAT} 
\end{subfigure}%
\begin{subfigure}[b]{0.25\textwidth}
   \includegraphics[scale=0.225]{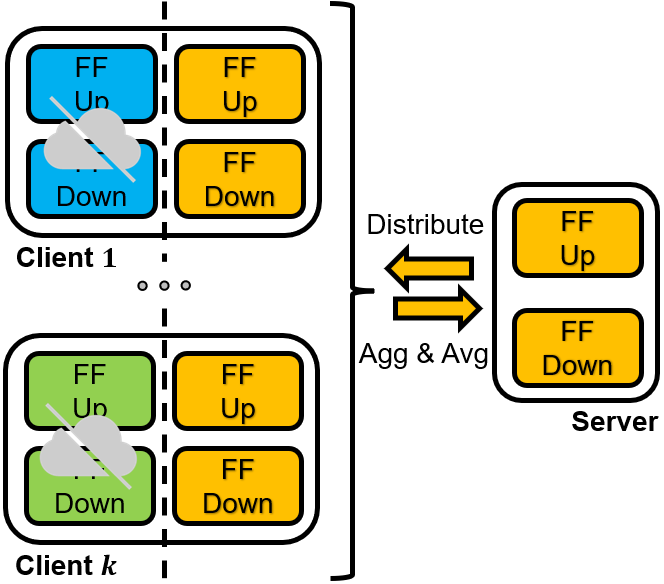} 
    \flushleft
    \caption{Communication}
    \label{fig:feddacom} 
\end{subfigure}%
\caption{Schematic illustration of the Dual-Adapter Teacher (\emph{DAT}) with local $A_c$ and frozen $\hat{A}_s$. Only the shared adapter $A_s$ is transmitted during federated communication.}
\vspace{-10pt}
\end{figure} 

\textbf{(1) Dual-Adapter Teacher (\emph{DAT}):} Before the first communication round, each client locally initializes the local adapter $A_c$ as well as the foundation model $f_{\theta}$ with the same pre-trained weights $\theta$. Subsequently, each client receives the parameters of $A_s$ from the server, which is then copied as $\hat{A}_s$ and kept frozen during the client local update. We combine $\hat{A}_s$ and $A_c$ as the Dual-Adapter Teacher (\emph{DAT}) and provide its schematic illustration in Figure \ref{fig:FedDAT}.

In \emph{DAT}, we constrain the parameters of $A_c$ strictly local for each client. By personalizing $A_c$, we force it to focus solely on client-specific knowledge, which is crucial for client data heterogeneity. Meanwhile, the frozen $\hat{A}_s$ is utilized to retain the client-agnostic knowledge captured by the shared adapter $A_s$. Similar to traditional adapters (Equation \ref{eq:adapter}), given the normalized output of FFN $h$ in a Transformer layer, \emph{DAT} performs the following transformation:
\begin{equation}
\small
h' \gets h + \frac{1}{2} \phi(h \cdot \hat{W}_s^{down}) \cdot \hat{W}_s^{up} + \frac{1}{2} \phi(h \cdot W_c^{down}) \cdot W_c^{up},
\end{equation}

where $\hat{W}_s$ and $W_c$ are the weight matrices for $\hat{A}_s$ and $A_c$, respectively. Afterwards, $T$ local update steps will be executed, in which the shared adapter $A_s$ and the \emph{DAT} module is optimized. 

By utilizing \emph{DAT} as a guidance for the local optimization of $A_s$ at each client, our goal is to distill client-specific knowledge into $A_s$ and mitigate the forgetting of $A_s$ on its client-agnostic knowledge. Hereby, we apply Mutual Knowledge Distillation (\emph{MKD}) for an efficient knowledge transfer, which will be introduced in the following. 

\begin{figure}[t]
\begin{subfigure}[b]{0.25\textwidth}
    \includegraphics[scale=0.28]{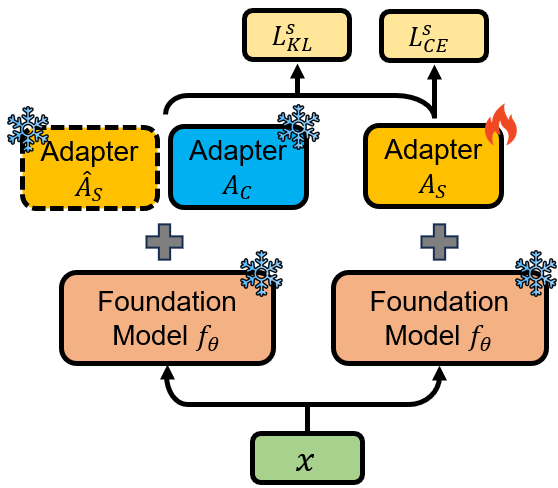} 
    \caption{Optimization of $A_s$}
\end{subfigure}%
\begin{subfigure}[b]{0.25\textwidth}
    \includegraphics[scale=0.28]{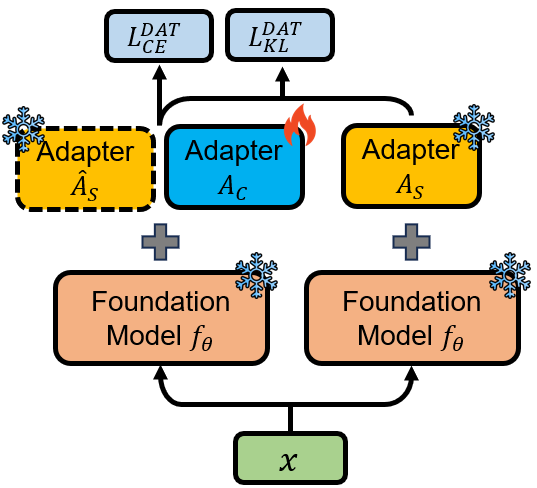} 
    \caption{Optimization of \emph{DAT}}
\end{subfigure}%
\caption{Schematic illustration of the Mutual Knowledge Distillation (\emph{MKD}) between \emph{DAT} and $A_s$.}
\label{fig:fedstage} 
\vspace{-10pt}
\end{figure}

\textbf{(2) Mutual Knowledge Distillation (\emph{MKD}):} A schematic illustration of \emph{MKD} is provided in Figure \ref{fig:fedstage}. \emph{MKD} executes bi-directional knowledge distillation between $A_s$ and \emph{DAT} via $L_{\text{KL}}^s$ and $L_{\text{KL}}^{\text{DAT}}$, respectively:

\vspace{-4pt}
\begin{equation}
\small
L_{\text{KL}}^s = \mathcal{KL}(z_s(x)||z_{\text{DAT}}(x)), \quad
L_{\text{KL}}^{\text{DAT}} = \mathcal{KL}(z_{\text{DAT}}(x)||z_s(x)),
\label{eq:as}
\vspace{4pt}
\end{equation}
 
where $\mathcal{KL}$ denotes the Kullback-Leibler divergence, $z_s$ and $z_{\text{DAT}}$ are the predicted logits of the foundation model injected with $A_s$ and \emph{DAT}, respectively. Hereby, this setup allows the shared adapter $A_s$ to capture both client-specific knowledge and client-agnostic stored in \emph{DAT} ($L_{\text{KL}}^s$). Additionally, we apply $A_s$ as guidance for the optimization \emph{DAT} ($L_{\text{KL}}^{\text{DAT}}$) to prevent possible overfitting, considering the scarce local data of each client \cite{mcmahan2017communication}. 

 
\emph{MKD} is utilized together with the guidance from ground-truth labels of the training data, i.e., 
\vspace{-4pt}
\begin{equation}
\small
\begin{aligned}
L_{\text{CE}}^s=&\sum_{c=1}^{C}\mathcal{I}(x, c) \cdot log ( \sigma(z_s(x))^{(c)} ), \\
L_{\text{CE}}^{\text{DAT}} = &\sum_{c=1}^{C}\mathcal{I}(x, c) \cdot log (\sigma(z_{\text{DAT}}(x))^{(c)}), 
\label{eq:ce}
\end{aligned}
\vspace{4pt}
\end{equation}

where, $\mathcal{I}(x, c)$ is a binary indicator (0 or 1) if $c$ is the ground-truth label for $x$, $\sigma$ is the softmax function. Hereby, we aim at training the foundation model, injected with either $A_s$ or \emph{DAT}, to correctly classify the training sample $x$. Finally, combining \emph{MKD} and $L_{CE}$ produces the optimization objective for $A_s$ and \emph{DAT}:

\begin{equation}
\small
\begin{aligned}
L^s=&L_{\text{CE}}^s + \alpha L_{\text{KL}}^s, \\
L^{\text{DAT}} = &L_{\text{CE}}^{\text{DAT}} + \beta L_{\text{KL}}^{\text{DAT}}, 
\label{eq:ce}
\end{aligned}
\end{equation}

where, $\alpha$ and $\beta$ are the weighting coefficient. While both \emph{DAT} and $A_s$ are randomly initialized, they become more informative as the training progresses. To reflect this observation, we apply an exponential ramp-up schedule for $\alpha$ and $\beta$. Despite the sophisticated design of our method, \texttt{FedDAT} indicates the same inference cost and communication overhead as the PEFT method $Adapter$, where only $A_s$ is transmitted and applied at deployment.

\section{Experiments and Analyses}

\begin{table*}[t]
\begin{center}
\small
\centering
\setlength\tabcolsep{7.5pt}
\begin{tabular}{c|cc|ccccc|c}

\toprule
Backbone & Method & \makecell[c]{Comm. Overhead} & VizWiz & COCO & Art & GQA & Abstract & Average\\
\hline
\multirow{8}{*}{ViLT} & $clf$-L & $-$ & 63.13{\scriptsize ±1.07} & 36.15{\scriptsize ±2.92} & 63.22{\scriptsize ±0.99} & 34.90{\scriptsize ±3.16} & 52.81{\scriptsize ±2.67} & 50.04{\scriptsize ±1.81}\\
~ & $LoRA$ & $0.60M (0.48\%)$ & 60.47{\scriptsize ±1.25} & 43.28{\scriptsize ±1.37} & 62.98{\scriptsize ±0.75} & 36.57{\scriptsize ±2.01} & 52.04{\scriptsize ±1.62} & 51.07{\scriptsize ±1.41} \\
~ & $prompt$ & $0.60M (0.48\%)$ & 60.13{\scriptsize ±1.05} & 52.13{\scriptsize ±0.87} & 63.02{\scriptsize ±1.58} & 39.09{\scriptsize ±0.37} & 52.88{\scriptsize ±3.07} & 53.45{\scriptsize ±2.04}\\
~ & $bias$ & $0.10M (0.08\%)$ & 61.83{\scriptsize ±2.41} & 49.41{\scriptsize ±2.36} & 69.38{\scriptsize ±1.69} & 40.43{\scriptsize ±0.66} & 60.36{\scriptsize ±1.92} & 56.28{\scriptsize ±1.97} \\
~ & $Adapter$-L & $-$ &  61.72{\scriptsize ±1.42} & 46.27{\scriptsize ±4.58} & 67.69{\scriptsize ±0.42} & 43.62{\scriptsize ±0.93} & 54.02{\scriptsize ±2.16} & 54.67{\scriptsize ±2.54}     \\
~ & $Adapter$ & $0.89M (0.75\%)$ & 61.39{\scriptsize ±1.11} & 52.39{\scriptsize ±6.20} & 68.72{\scriptsize ±3.20} & 43.72{\scriptsize ±0.65} & 59.43{\scriptsize ±2.94} & 57.13{\scriptsize ±4.08}  \\
~ & \textbf{FedDAT} & $0.89M (0.75\%)$ & 60.99{\scriptsize ±2.81} & 63.81{\scriptsize ±2.90} & 71.36{\scriptsize ±3.34} & 48.65{\scriptsize ±2.93} & 60.75{\scriptsize ±2.67} & \textbf{61.11}{\scriptsize ±2.98}    \\
\cline{2-9}
~ & $full$-L$^*$ & $-$ &  55.52{\scriptsize ±1.42} & 72.97{\scriptsize ±1.53} & 73.16{\scriptsize ±0.28} & 44.41{\scriptsize ±3.98} & 58.78{\scriptsize ±0.25} & 60.97{\scriptsize ±1.45} \\
~ & $full^*$& $87.40M (100\%)$ & 56.12{\scriptsize ±2.55} & 73.87{\scriptsize ±0.83} & 76.24{\scriptsize ±1.82} & 50.28{\scriptsize ±1.59} & 61.26{\scriptsize ±0.78} & 63.55{\scriptsize ±1.35}  \\

\hline
\multirow{8}{*}{VAuLT} 
~ & $clf$-L & $-$ & 61.83{\scriptsize ±1.85} & 32.42{\scriptsize ±0.04} & 64.52{\scriptsize ±1.55} & 35.08{\scriptsize ±5.57} & 48.48{\scriptsize ±0.77} & 48.46{\scriptsize ±1.15}\\
~ & $LoRA$ & $0.60M (0.29\%)$  & 62.17{\scriptsize ±1.32} & 40.56{\scriptsize ±0.86} & 63.08{\scriptsize ±1.13} & 33.47{\scriptsize ±3.08} & 47.34{\scriptsize ±1.04} & 49.32{\scriptsize ±1.16} \\
~ & $prompt$ & $0.60M (0.29\%)$ & 62.93{\scriptsize ±0.87} & 46.52{\scriptsize ±1.45} & 64.26{\scriptsize ±1.03} & 35.33{\scriptsize ±2.12} & 48.91{\scriptsize ±0.68} & 51.59{\scriptsize ±1.63} \\
~ & $bias$ & $0.21M (0.10\%)$ & 61.12{\scriptsize ±2.84} & 43.81{\scriptsize ±0.35} & 67.00{\scriptsize ±1.41} & 33.30{\scriptsize ±4.81} & 51.22{\scriptsize ±2.07} & 51.29{\scriptsize ±1.08}  \\
~ & $Adapter$-L & $-$ & 62.33{\scriptsize ±1.42} & 47.72{\scriptsize ±2.83} & 67.50{\scriptsize ±2.11} & 33.75{\scriptsize ±2.79} & 54.09{\scriptsize ±0.93} & 53.07{\scriptsize ±1.34}  \\
~ & $Adapter$ & $1.79M (0.77\%)$ & 52.53{\scriptsize ±3.65} & 53.63{\scriptsize ±0.28} & 66.80{\scriptsize ±0.53} & 35.65{\scriptsize ±1.84} & 50.03{\scriptsize ±1.77} & 51.73{\scriptsize ±0.46}    \\
~ & \textbf{FedDAT} & $1.79M (0.77\%)$ & 62.19{\scriptsize ±1.01} & 54.83{\scriptsize ±2.04} & 67.86{\scriptsize ±1.93} & 40.06{\scriptsize ±3.08} & 54.48{\scriptsize ±0.49} & \textbf{55.88}{\scriptsize ±1.79}     \\
\cline{2-9}
~ & $full$-L$^*$ & $-$ &  57.41{\scriptsize ±2.13} & 55.68{\scriptsize ±1.24} & 70.27{\scriptsize ±2.11} & 41.31{\scriptsize ±1.46} & 52.66{\scriptsize ±0.57} & 55.47{\scriptsize ±1.85}\\
~ & $full^*$& $227.77M (100\%)$ & 45.79{\scriptsize ±2.12} & 64.64{\scriptsize ±3.05} & 67.89{\scriptsize ±1.82} & 41.93{\scriptsize ±3.85} & 49.58{\scriptsize ±0.66} & 53.97{\scriptsize ±2.09}  \\

\hline
\multirow{7}{*}{ALBEF} & $LoRA$ & $1.52M (0.53\%)$ & 60.49{\scriptsize ±1.32} & 28.32{\scriptsize ±0.65} & 57.04{\scriptsize ±3.69} & 28.71{\scriptsize ±0.42}& 58.06{\scriptsize ±2.42}& 46.52{\scriptsize ±1.75}\\
~ & $prompt$ & $0.92M (0.32\%)$ & 63.13{\scriptsize ±0.65} & 32.50{\scriptsize ±1.20} & 63.45{\scriptsize ±0.42} & 32.08{\scriptsize ±1.07} & 59.45{\scriptsize ±1.78} & 50.12{\scriptsize ±0.95}\\
~ & $bias$ & $0.93M (0.32\%)$ & 63.23{\scriptsize ±0.14} & 31.23{\scriptsize ±0.28} & 61.23{\scriptsize ±1.12} & 35.93{\scriptsize ±1.73} & 57.88{\scriptsize ±0.28} & 49.90{\scriptsize ±0.87} \\
~ & $Adapter$-L & $-$ & 61.72{\scriptsize ±1.12} & 56.32{\scriptsize ±1.50} & 65.21{\scriptsize ±0.35} & 40.96{\scriptsize ±2.27} & 59.51{\scriptsize ±1.58} & 56.74{\scriptsize ±1.38}  \\
~ & $Adapter$ & $2.86M (0.98\%)$ & 59.52{\scriptsize ±2.44} & 69.35{\scriptsize ±2.78} & 68.32{\scriptsize ±0.89} & 41.02{\scriptsize ±3.12} & 60.83{\scriptsize ±2.66} & 59.81{\scriptsize ±1.87} \\
~ & \textbf{FedDAT} & $2.86M (0.98\%)$ & 61.52{\scriptsize ±1.51} & 76.36{\scriptsize ±0.63} & 71.04{\scriptsize ±0.50} & 49.22{\scriptsize ±1.60} & 63.65{\scriptsize ±1.19} & \textbf{64.36}{\scriptsize ±1.39}    \\
\cline{2-9}
~ & $full$-L$^*$ & $-$ & 61.22{\scriptsize ±0.14} & 77.80{\scriptsize ±1.39} & 74.45{\scriptsize ±0.7} & 50.09{\scriptsize ±1.06} & 63.58{\scriptsize ±2.79} & 65.43{\scriptsize ±1.37} \\
~ & $full^*$& $290.34M (100\%)$ & 51.91{\scriptsize ±1.42} & 78.38{\scriptsize ±1.11} & 75.65{\scriptsize ±0.14} & 55.91{\scriptsize ±0.54} & 70.47{\scriptsize ±0.83} & 66.46{\scriptsize ±0.96}  \\

\bottomrule       
\end{tabular}
\end{center}
\vspace{-7pt}
\caption{Evaluation results of different finetuning methods on our FL benchmark with distribution shift in both Vision and Language space. "L" indicates client local finetuning where no communication is involved. We report the mean{\scriptsize±std} accuracy of each client from 3 runs with different seeds.}
\label{tab:domainresults}
\vspace{-12pt}
\end{table*}

We conduct extensive empirical analyses to investigate the proposed method. Firstly, we compare \texttt{FedDAT} with other centralized PEFT methods on four heterogeneous FL benchmarks containing different Vision-Language tasks. Afterwards, we demonstrate the effectiveness of \texttt{FedDAT} components via ablation study. Finally, we analyze the promising convergence rate and scalability of \texttt{FedDAT}.

\subsection{Benchmark Experiments}

\subsubsection{Datasets Description:} We conduct experiments on different Vision-Language (VL) benchmarks with different types of data heterogeneity, including visual, textual, and task heterogeneity. We introduce these benchmarks in the following. 

\begin{figure}[b]
\centering
\vspace{-10pt}
\includegraphics[scale=0.2]{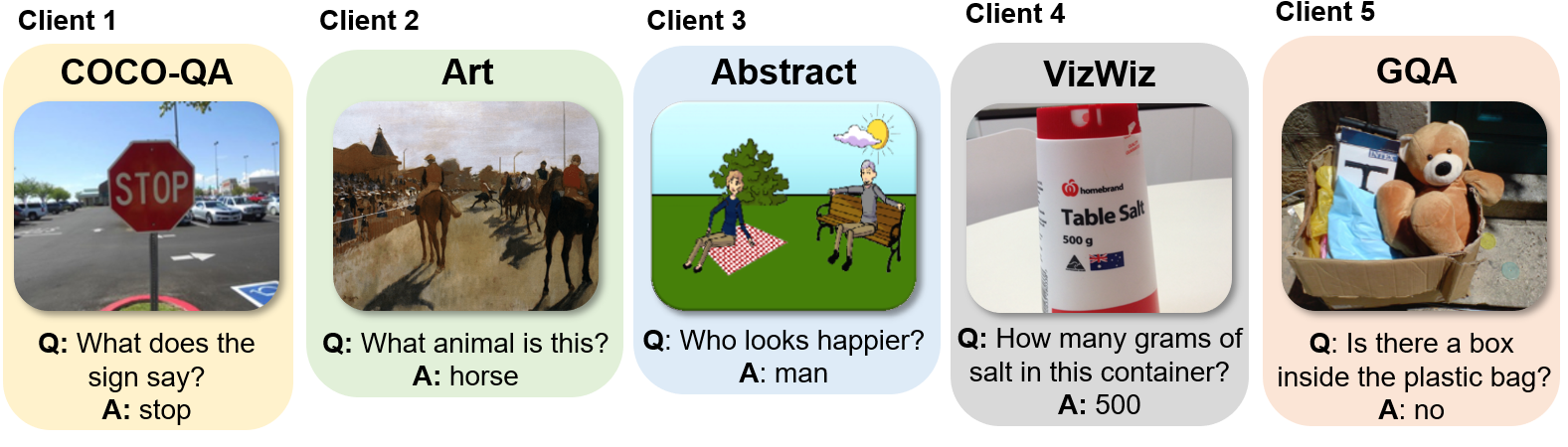} 
\vspace{-15pt}
\caption{Example VQA triplets of different datasets in \emph{Domain} benchmark with heterogeneity in both Vision and Language modality.}
\label{fig:domainex}
\vspace{-10pt}
\end{figure}

\begin{itemize}
    \item \textbf{Domain}: We adopt 5 common VQA datasets from different domains, i.e., VizWiz \cite{gurari2018vizwiz}, COCO QA \cite{ren2015exploring}, Art \cite{garcia2020AQUA}, GQA \cite{hudson2019gqa} and Abstract \cite{VQAabstract}. We assign one of the datasets to each client, leading to heterogeneity in both vision and language modality. Example VQA triplets from the benchmark are provided in Figure \ref{fig:domainex}.
    
    \item \textbf{Function} \& \textbf{Scene}: We adopt and split the CLOVE benchmark \cite{lei2023symbolic} into \emph{Scene} and \emph{Function} benchmark, which contains VQA triplets collected from 6 different visual environments and 5 different functions, respectively. Triplets from one scene (function) are allocated to one client, resulting in visual (textual) heterogeneity in the \emph{Scene} (\emph{Function}) benchmark. 

    \item \textbf{Task}: We adopt and modify the CLiMB benchmark \cite{srinivasan2022climb}, which contains 4 VL tasks, namely VQA \cite{VQAabstract}, Natural Language for Visual Reasoning (\emph{NLVR}) \cite{suhr2018corpus}, Visual Entailment (\emph{VE}) \cite{xie2019visual}, and Visual Commonsense Reasoning (\emph{VCR}) \cite{zellers2019vcr}. Each client owns data from one of the datasets, introducing task heterogeneity across different clients.
\end{itemize}

We downsample the original dataset to simulate client local data scarcity described in prior arts \cite{mcmahan2017communication} and provide more details in the Appendix. 

\subsubsection{Implementation Details:} For the task-heterogeneous benchmark (\emph{Task}), we adopt the Transformer encoder-only backbones following \cite{srinivasan2022climb}, i.e., ViLT \cite{kim2021vilt} and VAuLT \cite{chochlakis2022vault}. For the rest three benchmarks, we add another encoder-decoder backbone, i.e., ALBEF \cite{li2021align}. We compare \texttt{FedDAT} with various centralized PEFT methods adapted for FL, including $LoRA$ \cite{hu2021lora}, $prompt$-tuning \cite{guo2022promptfl}, and $bias$-tuning \cite{cai2020tinytl}. We also provide results of independent client optimization (marked by "$L$") of the classification head $clf$ and $Adapter$. Moreover, we provide the results of fully finetuning the models ($full$) as an \emph{oracle} method (marked by $*$), given the infeasibility of transmitting the entire foundation model in FL. 

\begin{algorithm}[t]
\small
\caption{Training procedure of FedDAT}\label{algo:FedDAT}
\textbf{ServerUpdate}
\begin{algorithmic}[1]
\STATE Randomly initialize $A_s$ \\
\FOR {round $r$ = 1 to $R$} 
    \FOR [\textbf{in parallel}]{client $k$ = 1 to $K$} 
        \STATE $A^k_s \gets$ ClientUpdate$(A_s, k, r)$
    \ENDFOR
    \STATE$A_s \gets \frac{1}{K} \sum_{k=1}^K  A^k_s$ \\
\ENDFOR
\vspace{0.5em}
\end{algorithmic} 
\textbf{ClientUpdate}$(A_s, k, r)$
\begin{algorithmic}[1]
\IF {$r=1$}
    \STATE Randomly initialize $A_c$ \\
\ENDIF
\STATE $\hat{A}_s \gets A_s$\\ 
\FOR {local step $t$ = 1 to $T$} 
    \STATE Sample $\{\boldsymbol{X}, \boldsymbol{y}\}$ from $D_k$ \\
    \STATE Optimize $A_s$ via minimizing $L^s$ \\ 
    \STATE Optimize \emph{DAT} via minimizing $L^{\text{DAT}}$ \\
\ENDFOR
\RETURN $A_s$
\end{algorithmic}
\end{algorithm}

To handle the different answer pools in different clients, we incorporate client-specific classification heads for ViLT and VAuLT, and apply client-specific answer lists for ALBEF. To make a fair comparison between different centralized PEFT algorithms and \texttt{FedDAT}, we apply the same hyperparameters search for all methods in different benchmarks. All experiments are repeated with 3 random seeds. The hyperparameters are detailed in the Appendix. 

\subsubsection{Results and Analyses:}
In Table \ref{tab:domainresults}, we provide the results of \texttt{FedDAT} and the other FL-adapted PEFT methods on our \emph{Domain} benchmark. We observe \texttt{FedDAT} outperforms all the baselines with all the architectures, achieving an average performance improvement of up to $4.55\%$ compared with the most promising baseline $Adapter$. This indicates the easy adaptability of \texttt{FedDAT} for both encoder-based and encoder-decoder-based VL models. Moreover, \texttt{FedDAT} depicts the same communication overhead as a single $Adapter$, which adds and optimizes only less than $1\%$ of the total parameters in the foundation model. This further illustrates its applicability to the FL system with constrained communication bandwidths. Besides, \texttt{FedDAT} narrows the performance gap between the PEFT methods and fully-finetuning methods. Interestingly, our approach outperforms the oracle methods $full$-L when applied on ViLT and VAuLT, which demonstrates the effectiveness of introducing client-specific knowledge into the client local optimization. We also note that applying $Adapter$-L for VAuLT, i.e., optimizing adapters for each client independently, achieves better results than $Adapter$, which provides additional evidence for our observation in Section \ref{sec:motivstudy}. 

Afterwards, we provide the comparison of clients' average accuracy between \texttt{FedDAT} and different PEFT methods on the other benchmarks. As shown in Table \ref{tab:funcsceneresults}, \texttt{FedDAT} provides promising improvements of up to $6.02\%$, $7.94\%$ and $1.09\%$ on \emph{Function, Scene}, and \emph{Task} benchmark, respectively. More details of the client specific performance are provided in the Appendix.

\begin{table}[t]
\begin{center}
\small
\centering
\setlength\tabcolsep{4.5pt}
\begin{tabular}{c|c|ccc}

\toprule 
Backbone & Method & Function & Scene & Task \\
\hline
\multirow{9}{*}{ViLT} 
~ & $clf$-L  & 31.58{\scriptsize ±1.97} & 24.52{\scriptsize ±0.95} & 49.46{\scriptsize ±0.39}\\
~ & $LoRA$  & 32.04{\scriptsize ±1.12} & 28.47{\scriptsize ±1.03} & 47.82{\scriptsize ±1.42}\\
~ & $prompt$ & 40.53{\scriptsize ±1.56} & 30.53{\scriptsize ±1.30} & 49.55{\scriptsize ±1.14} \\
~ & $bias$ & 43.81{\scriptsize ±1.39} & 33.65{\scriptsize ±1.87} & 50.71{\scriptsize ±1.26} \\
~ & $Adapter$-L  & 39.68{\scriptsize ±2.19} & 31.91{\scriptsize ±2.05} & 49.59{\scriptsize ±1.74} \\
~ & $Adapter$ & 48.37{\scriptsize ±1.56} & 31.07{\scriptsize ±1.08} & 51.44{\scriptsize ±1.34} \\
~ & \textbf{FedDAT} & \textbf{54.39}{\scriptsize ±2.36} & \textbf{39.35}{\scriptsize ±1.25} & \textbf{52.37}{\scriptsize ±0.52} \\
\cline{2-5}
~ & $full$-L$^*$ & 56.81{\scriptsize ±2.97} & 38.00{\scriptsize ±1.48} & 50.64{\scriptsize ±1.42} \\
~ & $full^*$ & 59.62{\scriptsize ±2.56} & 40.62{\scriptsize ±3.76} & 53.17{\scriptsize ±0.69} \\
    
\hline
\multirow{8}{*}{VAuLT} & $clf$-L & 27.72{\scriptsize ±3.05} & 21.22{\scriptsize ±2.08} & 39.63{\scriptsize ±1.07} \\
~ & $LoRA$ & 29.87{\scriptsize ±1.86} & 23.08{\scriptsize ±1.09} & 38.35{\scriptsize ±1.47} \\
~ & $prompt$ & 36.32{\scriptsize ±2.07} & 25.63{\scriptsize ±1.54} & 38.75{\scriptsize ±1.34}\\
~ & $bias$ & 36.11{\scriptsize ±3.05} & 24.89{\scriptsize ±2.17} & 39.46{\scriptsize ±0.99} \\
~ & $Adapter$-L & 37.22{\scriptsize ±2.38} & 28.57{\scriptsize ±1.98} & 40.42{\scriptsize ±1.21} \\
~ & $Adapter$ &  41.50{\scriptsize ±3.24} & 29.39{\scriptsize ±2.65} & 40.19{\scriptsize ±0.89} \\
~ & \textbf{FedDAT} & \textbf{44.54}{\scriptsize ±2.08} & \textbf{34.31}{\scriptsize ±2.87} & \textbf{41.28}{\scriptsize ±0.57} \\
\cline{2-5}
~ & $full$-L$^*$  & 49.13{\scriptsize ±2.68}  & 35.11{\scriptsize ±1.99} & 41.66{\scriptsize ±1.32} \\
~ & $full^*$ & 46.38{\scriptsize ±1.57} & 36.72{\scriptsize ±2.57} & 42.44{\scriptsize ±0.71} \\


\bottomrule       
\end{tabular}
\end{center}
\vspace{-7pt}
\caption{Evaluation results of different methods on \emph{Function}, \emph{Scene}, and \emph{Task} benchmark. "L" indicates independent client finetuning. We report the mean{\scriptsize±std} accuracy of 3 trials.}
\label{tab:funcsceneresults}
\vspace{-12pt}
\end{table}

\subsection{Ablation Study}

\begin{table}[b]
\begin{center}
\small
\centering
\setlength\tabcolsep{3pt}
\begin{tabular}{c|c|ccc}

\toprule
Stage & Method & Domain & Function & Scene \\

\hline
- & $Adapter$ & 57.13{\scriptsize ±4.08} & 48.37{\scriptsize ±1.56} & 31.07{\scriptsize ±1.08} \\
\hline
\multirow{4}{*}{Optimization} & {\scriptsize w/o} $\hat{A}_s$ & 58.24{\scriptsize ±0.98} & 50.62{\scriptsize ±1.45} & 33.04{\scriptsize ±0.65} \\
~ & {\scriptsize w/o} $A_c$ & 57.87{\scriptsize ±1.24} & 50.93{\scriptsize ±0.85} & 32.45{\scriptsize ±0.27} \\
~ & {\scriptsize w/o} \emph{MKD} & 58.41{\scriptsize ±1.57} & 52.82{\scriptsize ±2.98} & 36.98{\scriptsize ±1.07} \\
~ & \textbf{FedDAT}  & \textbf{61.11}{\scriptsize ±2.98} & \textbf{54.39}{\scriptsize ±2.36} & \textbf{39.35}{\scriptsize ±1.25} \\
\hline
\multirow{3}{*}{Inference} & $A_c + A_s$ & 58.45{\scriptsize ±1.57} & 50.42{\scriptsize ±1.87} & 35.61{\scriptsize ±2.41} \\
~ & $A_c$ & 55.87{\scriptsize ±3.35} & 46.14{\scriptsize ±2.60} & 32.84{\scriptsize ±0.78}  \\
~ & $A_s$ (\textbf{FedDAT})  & \textbf{61.11}{\scriptsize ±2.98} & \textbf{54.39}{\scriptsize ±2.36} & \textbf{39.35}{\scriptsize ±1.25} \\

\bottomrule       
\end{tabular}
\end{center}
\caption{Ablation study for different components in optimization and inference stage of \texttt{FedDAT} on three benchmark datasets. }
\label{tab:ablation}
\vspace{-10pt}
\end{table}

\begin{figure*}[t]
    \includegraphics[scale=0.345]{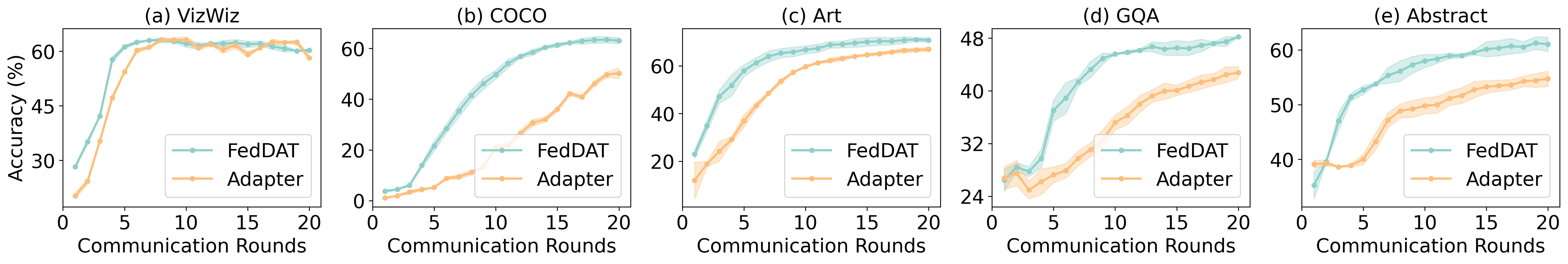} 
    \flushleft
    \vspace{-7pt}
    \caption{Convergence analysis of ViLT model on different clients in \emph{Domain} benchmark.} 
    \vspace{-10pt}
    \label{fig:converg}
\end{figure*}

To illustrate the importance of different components used in \texttt{FedDAT}, we conduct an ablation study for ViLT on three benchmarks. The results are shown in Table \ref{tab:ablation}. We first investigate the optimization process, where we notice that optimizing without \emph{DAT}, i.e., applying solely the local adapter $A_c$ or the frozen adapter $\hat{A_s}$ as the teacher, leads to only minimal performance increase, which indicates the effectiveness of our Dual-Adapter Teacher design. Besides, distilling only the knowledge from \emph{DAT} to the shared adapter $A_s$, i.e., omitting the bi-directional \emph{MKD}, brings visible performance gain. Combining both strategies achieves the best results, which further demonstrates their complementarity. Additionally, we validate other inference choices. Specifically, we evaluate the final \emph{DAT} module (combination of $A_c$ and $\hat{A}_s$) and the local adapter $A_c$ at each client. We again note that we are addressing the problem of finetuning a global foundation model via FL, where no further personalization is required. Considering the inference efficiency and the problem setting, we adopt the shared adapter $A_s$ for inference, which also achieves the most promising results. 

\subsection{Convergence Analysis}
In Figure \ref{fig:converg}, we display the convergence analysis of \texttt{FedDAT} compared with the most promising PEFT method $Adapter$ on \emph{Domain} benchmark. Hereby, we report the accuracy of the clients on their corresponding local testing set after each communication round. As shown in the figure, even though \texttt{FedDAT} utilizes a more sophisticated optimization schema, i.e., a combination of \emph{DAT} and \emph{MKD}, the learning curves of \texttt{FedDAT} still exhibit faster convergence rates than single $Adapter$. It is also worth noticing that \texttt{FedDAT} already achieves distinct performance gain after 5 communication rounds, i.e., $25\%$ of the total communication budgets.

\subsection{Scalability Analysis of \texttt{FedDAT}}
To show the effectiveness of \texttt{FedDAT} applied on ViLT and ALBEF under various application scenarios, we further conduct experiments on \emph{Function} and \emph{Scene} benchmarks with different numbers of clients. More specifically, we split the data of each function in the original CLOVE dataset \cite{lei2023symbolic} into 5 subsets, where each subset has an equal number of training data and is assigned to one client, following the client data scarcity described in \cite{mcmahan2017communication}. We conduct experiments where 1, 2, 3, 4, and 5 clients (subsets) from each function are selected, which gives in total 5, 10, 15, 20, and 25 clients joining the federated communication for the \emph{Function} benchmark, respectively. We apply also the same split strategy for the 6 different visual environments for the \emph{Scene} benchmark and conduct the same experiment. We utilize a larger communication budget when more clients are joining FL, since more clients correspond to a larger data quantity. More details regarding the experimental setups are provided in Appendix.

We observe that \texttt{FedDAT} consistently outperforms $Adapter$ across all setups with small or large quantities of training data. Notably, a performance gap of up to $10\%$ for ALBEF and $6\%$ for ViLT is evident. These results indicate the scalability of \texttt{FedDAT} in handling complex FL applications involving a larger number of clients and increased communication budgets. Furthermore, we note that ALBEF exhibits a performance increase when more clients join FL, while ViLT indicates the opposite trend. We attribute this difference to the possible limitation in the model capacity of ViLT, leading to performance degradation when handling a larger client population. 

\begin{figure}[H]
\includegraphics[scale=0.255]{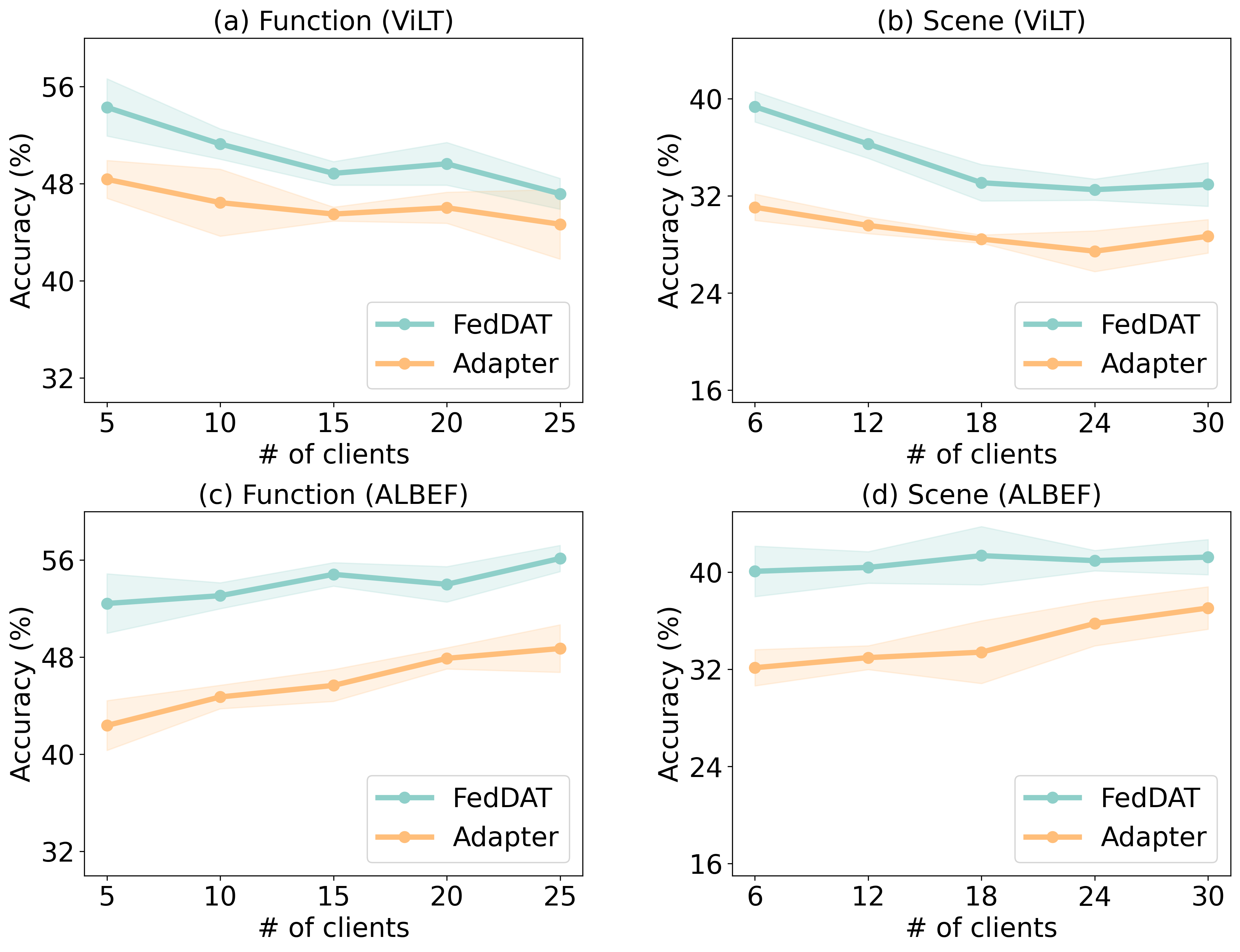} 
\centering
\vspace{-15pt}
\caption{Scalability analysis of \texttt{FedDAT} with different number of clients on \emph{Funciton} and \emph{Scene} benchmarks.}
\label{fig:fedd} 
\end{figure}

\section{Conclusion}
In this work, we propose the first FL framework to address the parameter-efficient finetuning (PEFT) of the foundation model in heterogeneous FL, where various Vision-Language tasks are investigated. The proposed method, named \texttt{FedDAT}, optimizes a shared adapter utilizing the Dual-Adapter Teacher (\emph{DAT}) and Mutual Knowledge Distillation (\emph{MKD}). Specifically, \emph{DAT} comprises a copy of the shared adapter, kept frozen, and a strictly local adapter at each client. This design allows it to retain both client-agnostic knowledge and client-specific knowledge. Afterwards, \emph{MKD} is applied for an efficient bi-directional knowledge transfer between \emph{DAT} and the shared adapter. Consequently, both types of knowledge are distilled from \emph{DAT} into the shared adapter, while the shared adapter could mitigate the local potential overfitting of \emph{DAT}. Compared with existing centralized PEFT methods, \texttt{FedDAT} achieves promising results on the four FL benchmarks with various Vision-Language tasks, demonstrating its effectiveness. Additional experiments indicate its applicability to complex FL setups involving larger distributed systems and increased training budgets.

\bibliography{main}
\bibliographystyle{acl_natbib}
\end{document}